%% file: main.tex
\documentclass[10pt,twocolumn,letterpaper]{article}

\usepackage{cvpr}              

\usepackage{graphicx}
\usepackage{amsmath}
\usepackage{amssymb}
\usepackage{booktabs}
\usepackage{bm}
\usepackage{soul}
\usepackage{multicol}
\usepackage{multirow}
\usepackage{amsmath}
\usepackage{tabularx}
\usepackage[accsupp]{axessibility}  

\usepackage{color}

\newlength\savewidth\newcommand\shline{\noalign{\global\savewidth\arrayrulewidth
  \global\arrayrulewidth 1pt}\hline\noalign{\global\arrayrulewidth\savewidth}}

\usepackage[pagebackref,breaklinks,colorlinks]{hyperref}

\usepackage[capitalize]{cleveref}
\crefname{section}{Sec.}{Secs.}
\Crefname{section}{Section}{Sections}
\Crefname{table}{Table}{Tables}
\crefname{table}{Tab.}{Tabs.}

\def\cvprPaperID{3841} 
\def\confName{CVPR}
\def\confYear{2022}
\renewcommand\footnotemark{}

\begin{document}

\title{GPV-Pose: Category-level Object Pose Estimation via Geometry-guided Point-wise Voting}

\author{Yan Di$^{1*}$, Ruida Zhang$^{2*}$, Zhiqiang Lou$^{2}$, Fabian Manhardt$^{3}$,\\
Xiangyang Ji$^{2}$, Nassir Navab$^{1}$ and Federico Tombari$^{1,3}$\\
\textsuperscript{1}Technical University of Munich, \textsuperscript{2}Tsinghua University, \textsuperscript{3}Google\\
\tt\small{\{\textsuperscript{*}zhangrd21@mails. lzq20@mails. xyji@\}tsinghua.edu.cn},
\tt\small{shangbuhuan13@gmail.com}
\\
\tt\small{fabianmanhardt@google.com}, \tt\small{tombari@in.tum.de}
\thanks{\textsuperscript{*}Authors with equal contributions.}
\thanks{Codes: \url{https://github.com/lolrudy/GPV_Pose}}
}
\maketitle

\begin{abstract}
While 6D object pose estimation has recently made a huge leap forward, most methods can still only handle a single or a handful of different objects, which limits their applications. 
To circumvent this problem, category-level object pose estimation has recently been revamped, which aims at predicting the 6D pose as well as the 3D metric size for previously unseen instances from a given set of object classes.
This is, however, a much more challenging task due to severe intra-class shape variations.
To address this issue, we propose GPV-Pose, a novel framework for robust category-level pose estimation, harnessing geometric insights to enhance the learning of category-level pose-sensitive features. 
First, we introduce a decoupled confidence-driven rotation representation, which allows geometry-aware recovery of the associated rotation matrix.
Second, we propose a novel geometry-guided point-wise voting paradigm for robust retrieval of the 3D
object bounding box.
Finally, leveraging these different output streams, we can enforce several geometric consistency terms, further increasing performance, especially for non-symmetric categories.
GPV-Pose produces superior results to state-of-the-art competitors on common public benchmarks, whilst almost achieving real-time inference speed at 20 FPS. 
\end{abstract}

\input{sections/intro1115}
\input{sections/relatedworks1115}
\input{sections/method1115}

\input{sections/experiments1115}

\input{sections/conclusion1022}
{\small
\bibliographystyle{ieee_fullname}
\bibliography{egbib}
}

\end{document}

%% file: sections/intro1115.tex
\section{Introduction}

\begin{figure}[t]
  \centering
  \includegraphics[width=0.99\linewidth]{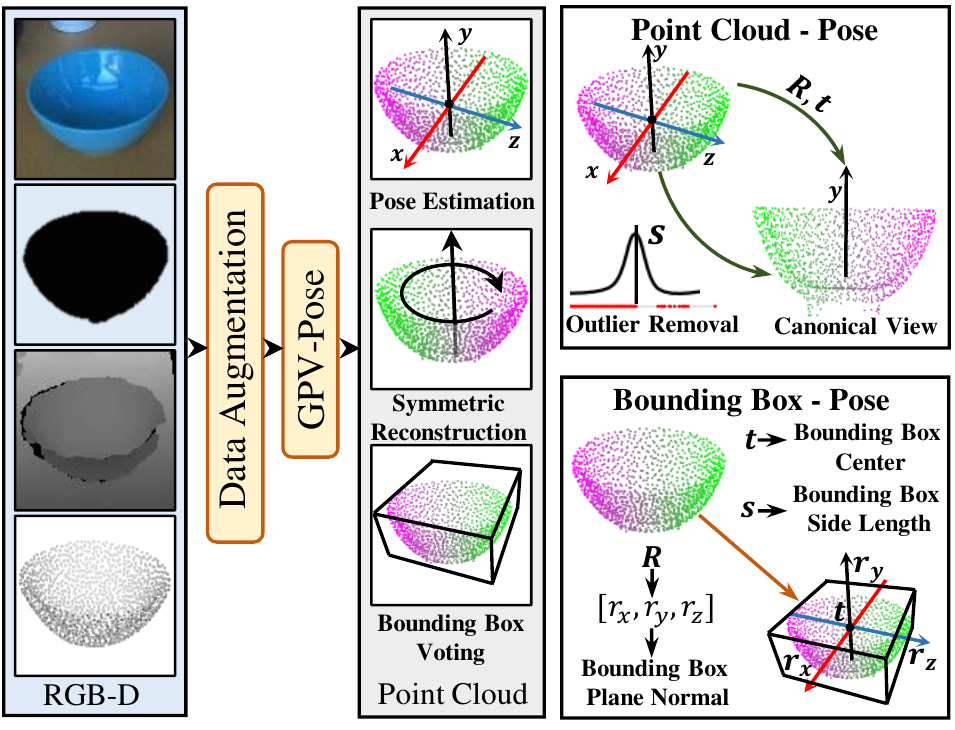}
  \caption{\textbf{GPV-Pose} consists of three individual branches for pose estimation, symmetry-aware reconstruction and point-wise bounding box voting.
  To enhance the learning of pose-sensitive features, we make use of two different types of geometric relations, \ie \textbf{Point Cloud - Pose} (\textbf{PP}) and \textbf{Point Cloud - Bounding Box - Pose} (\textbf{PBP}). 
  In the \textbf{PP} stream, we explicitly analyze how to directly retrieve pose from the point cloud, while in the \textbf{PBP} stream, we first introduce a geometry-guided point-wise bounding box voting approach and then establish geometric correspondences  between pose and the bounding box.
  }
  \label{teaser}
\end{figure}

Being able to reliably estimate the 6D object pose, \ie its 3D rotation and translation, is a fundamental problem in computer vision, as it enables a wide range of applications within AR\&VR~\cite{arvr}, robotics~\cite{robotics} and 3D understanding~\cite{nie2020total3dunderstanding, zhang2021holistic}. 
Hence, a lot of research has recently been devoted to the domain of pose estimation, producing methods that can reliably estimate the pose at real-time, even under severe occlusion~\cite{hybridpose,densefusion,GDRN,sopose}. 
Nevertheless, the majority of these methods can only deal with a few objects, in fact, sometimes even just a single instance at a time~\cite{labbe2020cosypose,Kehl2017}.
In addition, a high-quality CAD model is usually required during training and/or inference~\cite{Kehl2017,zakharov2019dpod}, which clearly limits the use of such methods in real applications. 
To deal with this issue, category-level pose estimation attempts to go beyond the instance-level scenario and estimate the pose together with the object's scale for previously unseen objects from known classes~\cite{NOCS,cass}. 
The category-level task is inherently more challenging due to the lack of respective CAD models as well as the large intra-class variations among different objects. 

Despite category-level pose estimation being a well-established field~\cite{lopez2011deformable,ozuysal2009pose, savarese20073d}, it has very recently started to enjoy again increasing popularity, thanks to a new line of works based on deep learning~\cite{NOCS,cass,shape_deform}. 
Interestingly, most of these works resort to a learned or manually-designed canonical object space to recover the pose~\cite{NOCS, cass} and, in parts, additionally leverage point-cloud based shape priors~\cite{shape_deform,sgpa} or pose consistency terms~\cite{dualposenet} to better deal with intra-class shape variations. Nonetheless, despite these methods achieving remarkable improvements on the benchmarks, their performance is still far from satisfactory due to their insufficient ability to extract pose-sensitive features, as they do not explicitly harness the geometric relationships between pose and point cloud.

In this paper, we propose GPV-Pose, a novel category-level pose estimation framework that leverages geometric constraints to enhance the learning of intra-class object shape characteristics.
GPV-Pose stacks three individual branches on top of a 3D graph convolution (3DGC) based encoder~\cite{3DGC, fs-net} for direct pose regression, symmetry-aware reconstruction and point-wise bounding box voting. 
In particular, we introduce a novel decoupled confidence-driven rotation representation, in which the rotation matrix is decomposed as the plane normals of the object bounding box.
We demonstrate that our estimated confidence enables closed-form recovery of the 3D rotation and is capable of capturing geometric characteristics of each class. 
Further, besides enhancing the feature quality by means of symmetry-aware reconstruction, we additionally propose a novel \textbf{G}eometry-guided \textbf{P}oint-wise \textbf{V}oting paradigm (GPV), enabling the robust recovery of the object's 3D bounding box.
As constituted in Fig.~\ref{teaser}, leveraging these three individual estimates, two streams of geometric relationships, \ie \textbf{Point Cloud - Pose} (\textbf{PP}) and \textbf{Point Cloud - Bounding Box - Pose} (\textbf{PBP}), can be exploited to serve as geometric consistency terms, further improving the learning of category-level shape characteristics and in turn enhancing the overall performance.

In summary, our main contributions are as follows:
\begin{itemize}
\setlength{\itemsep}{0pt}
\setlength{\parsep}{0pt}
\setlength{\parskip}{0pt}
\item 
We propose a novel geometry-guided category-level pose estimation framework GPV-Pose, which consists of three respective branches for direct pose regression, symmetry-aware reconstruction, and point-wise bounding box voting, giving superior results on common public benchmarks at a high framerate of 20 FPS.
\item 
To enhance the learning of intra-class shape characteristics, we introduce confidence-aware predictions of rotation and the object bounding box, and two parallel streams of geometric constraints, \textbf{PP} and \textbf{PBP}, are naturally converted into geometric consistency terms.
\item 
We propose a novel point-wise object bounding box voting mechanism that aggregates direction, distance and confidence predictions of all points with the confidence-weighted least square algorithm.
\end{itemize}

%% file: sections/relatedworks1115.tex
\section{Related Works}
\textbf{Instance-level 6D Pose Estimation.}
Instance-level pose estimation describes the task of estimating 6D pose for known object instances. 
For monocular methods, there are three different lines of works. 
Whereas a few methods directly estimate the pose~\cite{Kehl2017, xiang2017posecnn, manhardt2018deep, manhardt2019explaining, li2019deepim, labbe2020cosypose}, other approaches either learn a latent embedding for latter pose retrieval~\cite{Sundermeyer_2018_ECCV,sundermeyer2020multi} or predict 2D-3D correspondence and use a variant of the P\textit{n}P paradigm to obtain the final pose~\cite{li2019cdpn,hybridpose,peng2019pvnet,zakharov2019dpod,hodan2020epos,park2019pix2pose}. 
Interestingly, a handful methods combine direct regression with correspondence-driven methods, in particular, they first estimate 2D-3D correspondences and then make use of another network to learn the P\textit{n}P step~\cite{hu2020single, sopose, GDRN}.
A similar division can be also made for RGB-D methods. While a few methods again learn a latent embedding for retrieval purposes~\cite{wohlhart2015learning, Kehl2016a}, most methods aim at directly estimating the final 6D object pose~\cite{6drgbd, wang2019densefusion, he2020pvn3d, FFB6D}. The main difference lies thereby in the way how the two different modalities are fused together. 

Despite great progress in the recent years, instance-level methods can typically only deal with a single or a handful of objects and require an object CAD model for training and testing, limiting its use in practical applications such as autonomous driving.

\textbf{Category-level Pose Estimation.}
Category-level approaches aim at predicting the pose of previously unseen objects~\cite{manhardt2020cps,NOCS}.
Exemplary, Sahin \etal~\cite{9d1} introduce \textit{"Intrinsic Structure Adaptor"} to adapt the distribution shifts arising from shape discrepancies.
Wang~\etal~\cite{NOCS} derive a \textit{Normalized Object Coordinate Space} (NOCS) and recover pose using the Umeyama's algorithm~\cite{umeyama1991least}, while CASS~\cite{cass} introduces a learned canonical shape space.
6D-PACK~\cite{6dpack} computes the pose by means of tracking.
To alleviate the influence of intra-class shape variations, a few methods incorporate point-based object shape priors~\cite{shape_deform, sgpa, donet}.
DualPoseNet~\cite{dualposenet} instead utilizes a dual network for explicit and implicit pose prediction, and introduces a pose refinement strategy by means of pose consistency within both branches.
While FS-Net~\cite{fs-net} proposes to represent rotation using decoupled vectors, DO-Net~\cite{donet} makes use of symmetry for pose optimization.
Additionally, SGPA~\cite{sgpa} adopt visual transformers~\cite{transformer} for pose estimation.
Interestingly, a handful of methods have recently started to investigate pose estimation for articulated objects~\cite{aticategory, captra}.
Noteworthy, none of the aforementioned works harness different geometric cues to strengthen the model's prediction capabilities, which in turns mitigates the overall performance. 

On the contrary, in this paper we propose a novel geometry-guided category-level pose estimation framework, leveraging two different streams of geometric information based on the sampled point cloud and the object bounding box, which we refer to as \textbf{PP} and \textbf{PBP}.

%% file: sections/method1115.tex
\begin{figure*}[h]
  \centering
  \includegraphics[width=0.99\linewidth]{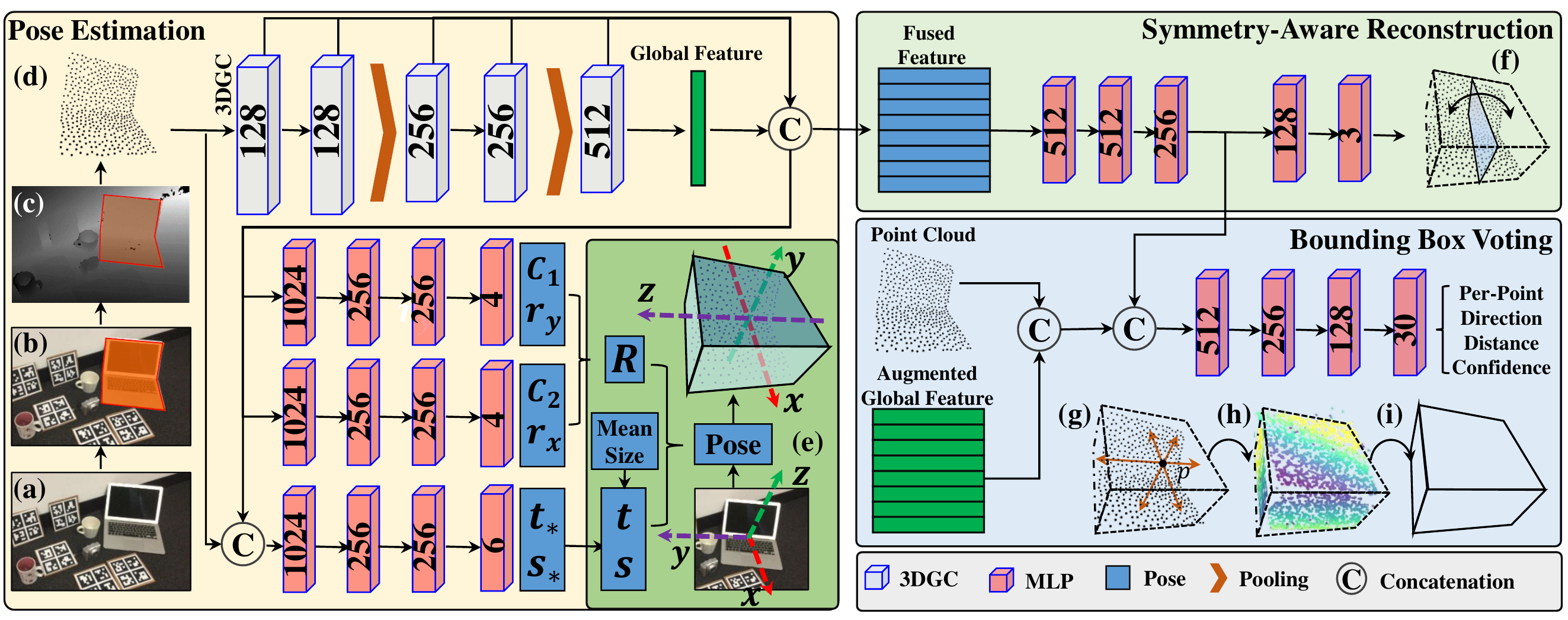}
  \vspace{-0.3cm}
  \caption{\textbf{Schematic overview of our architecture for GPV-Pose.} 
  We first employ an off-the-shelf object detector (\eg Mask-RCNN~\cite{maskrcnn}) to segment the object of interest from the depth map, and sample 1028 points from the backprojected depth map as input to GPV-Pose (a-d).
  We then extract global and per-point features with the help of 3DGC~\cite{3DGC}, processed with three respective branches.
  The first branch (yellow) outputs confidence-aware rotation $\{\bm{c_1, r_y, c_2, r_x}\}$, residual translation $\bm{t_{*}}$ and size $\bm{s_{*}}$ parameters, from which we recover the pose $\{\bm{R, t, s}\}$ in closed form.
  Moreover, while the second branch (green) reconstructs the input in a symmetry-aware manner (f), the last branch (blue) conducts point-wise bounding box voting.
  For each point in the point cloud (g), we predict its direction, distance and confidence towards each bounding box face, moving the points to the respective bounding box faces (h).
  By employing confidence-weighted least squares we can recover the plane parameters (i). Both branches are then further leveraged to serve as additional geometric guidance during optimization.
  Note that, at inference, we only need the pose estimation part, with benefits in efficiency.
  }
  \label{pipeline}
\end{figure*}

\section{GPV-Pose}
Given a RGB-D image, we first employ an off-the-shelf object detector (\eg Mask-RCNN~\cite{maskrcnn}) to segment the object of interest from the depth map.
We then sample 1028 points from the backprojected 3D point cloud and feed them as input to our proposed GPV-Pose network.
Since 3DGC~\cite{3DGC} is insensitive to shift and scale of the given point cloud, we employ 3DGC as the backbone to extract global and per-point features.
We further attach three parallel branches for direct confidence-aware pose prediction, symmetry-aware point cloud reconstruction and point-wise bounding box voting.
Leveraging these different outputs, we can naturally enforce several geometric constraints, leading to superior performance. A schematic overview is provided in Fig~\ref{pipeline}.
In the following subsections, we will explain in detail each branch as well as our geometric constraints. 
Note that throughout this paper we use $\{\bm{R, t, s}\}$ to refer to $\{Rotation, Translation, Size\}$.

\subsection{Confidence-aware Pose Regression}
Recent works~\cite{zhou2019continuity, fs-net} have shown that predicting the rotation in the form of bounding box plane normals, \ie equivalent to considering two columns of the rotation matrix, can benefit learning as it resolves discontinuities in SO(3).
Hence following these works, as shown in Fig.~\ref{pipeline} (e), we also predict rotation as two plane normals of the 3D bounding box.
However, since certain normals are naturally easier to recover, we additionally estimate a confidence value for each normal in an effort to increase robustness when recovering the final 3D rotation.
As an example, imagine predicting the rotation for a \textit{laptop}, as one plane normal is typically perpendicular to the keyboard surface, it is thus an easier target to compute than the remaining one.
Therefore, this normal should receive higher confidence to, in turn, stabilize rotation recovery.
Overall, we want to enforce that a higher confidence is accompanied by a more accurate rotation normal prediction. 
Thus, we define,
\begin{equation}
    \mathcal{L}^{Basic}_{r_c}=\lambda_1 \sum_{i \in \{x, y\}} \left \| c_{i} - exp(-k_1|r_i - r_i^{gt}|^2) \right \|_1,
    \label{basic_c}
\end{equation}
where $k_1$ is a constant, $r_i^{gt}$ is the corresponding ground truth plane normal and
$\left\|\cdot \right\|_1$ denotes the $\mathcal{L}_1$-loss.

As shown in Fig.~\ref{voteandR} (b), given predicted plane normals $r_y$, $r_x$, and their confidence $c_y$, $c_x$, we minimize the following cost function to calibrate the plane normals to be perpendicular normals $r_{y'}$ and $r_{x'}$,
\begin{equation}
\begin{split}
\theta_1^*, \theta_2^* &= arg \;\; min \;\; c_y\theta_1^2 + c_x\theta_2^2  \\
& s.t. \;\; \theta_1 + \theta_2 + \pi/2 = \theta,
\end{split}
\label{minimum}
\end{equation}
where $\theta$ denotes the angle between $r_x$ and $r_y$,
From Eq.~\ref{minimum} we then obtain
\begin{equation}
\left\{\begin{matrix}
\theta_1^* = \frac{c_x}{c_x+c_y}(\theta-\frac{\pi }{2})\\ 
\theta_2^* = \frac{c_y}{c_x+c_y}(\theta-\frac{\pi }{2}).
\end{matrix}\right.
\end{equation}

\noindent 
The calibrated plane normals $r'_y$ and $r'_x$ can be calculated from $(\theta_1^*,\theta_2^*)$ using the Rodrigues Rotation Formula.
Eventually, the rotation matrix is obtained as $\bm{R'}=[r'_x, r'_y, r'_x \times r'_y ]$.
As for the translation $\bm{t}$, provided the predicted residual translation $\bm{t_*}$ and the mean of the input point cloud $M_P$, we compute $\bm{t} = \bm{t_*} + M_P$.
Similarly, given the estimated residual size $\bm{s_*}$ and the pre-computed category mean size $C_m$, we obtain the scale $\bm{s} = \bm{s_*} + C_m$.
Please refer to the supplementary material for all derivations.

\subsection{Symmetry Analysis}
We utilize two types of symmetries to extract more effective per-point features: \textbf{Reflection Symmetry} and \textbf{Rotational Symmetry} following~\cite{donet}.
For reflection symmetry categories (\textit{mug}, \textit{laptop}), we predict the corresponding point cloud $P'$ of the input point cloud $P$ w.r.t. the reflection plane, while for rotational symmetry categories (\textit{can, bowl, bottle}), we predict $P'$ that is symmetrical to $P$  w.r.t the symmetry-axis, otherwise we directly reconstruct $P$ with $P'=P$.
The symmetry-aware reconstruction is supervised by the following loss term
\begin{equation}
\mathcal{L}^{Basic}_{(sym)} = \lambda_2 \left\|P'- \varepsilon(P, R, t, R_{gt}, t_{gt})\right\|_1,
    \label{sym_loss}
\end{equation}
where $\varepsilon(*)$ depends on the symmetry type.
Due to space limit, please refer to the supplementary material and~\cite{donet} for detailed derivations.


\subsection{Point-wise Voting for Bounding Box}
\label{BBoxVote}
\begin{figure}[t]
  \centering
  \includegraphics[width=0.9\linewidth]{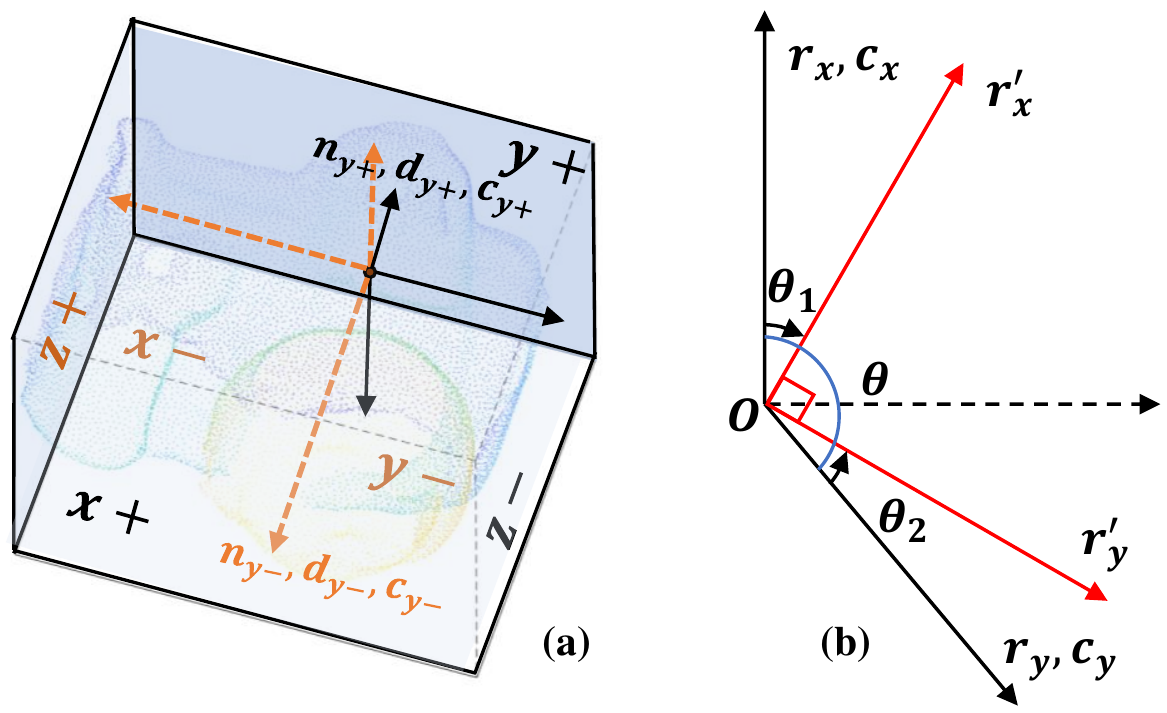}
  \caption{Confidence-aware predictions of the bounding box (a) and rotation (b).
  In (a), for each point, we predict its direction $n$, distance $d$ and confidence $c$ to each of the six faces, \eg for face $y+$, we predict the $n_{y+},d_{y+},c_{y+}$ of each point.
  In (b), we calibrate the predicted rotation normals $r_x, r_y$ to be perpendicular normals $r_{x'},r_{y'}$ by minimizing  Eq.~\ref{minimum}.
   }
  \label{voteandR}
\end{figure}

Another major contribution of GPV-Pose resides in our novel confidence-weighted point-wise voting strategy for robust bounding box prediction.
To this end, for each observed point $p_j$, we predict its direction ${n^j_i}$, distance $d^j_i$ and confidence $c^j_i$ towards each of the six bounding box faces with $i \in \mathcal{B}$ and $\mathcal{B}=\{y\pm, x\pm, z\pm\}$, as shown in Fig.~\ref{voteandR} (a).
We then compute the bounding box of the observed object using weighted averaging of all spatial cues.
As depicted in Fig.~\ref{voteandR} (a), consider the top bounding box face $y+$, we obtain the corresponding point $p'_j$ on the top face $y+$ for $p_j$ as
\begin{equation}
    p'_j = p_j + {n^j_{y+}}d^j_{y+}.
\end{equation}
\noindent The plane parameters, described as the normal $N_{y+}$ and origin-to-plane distance $D_{y+}$ for $y+$, can be estimated using weighted least squares, where the weight of point $p_j$ amounts to its confidence $c_j$. 
All other faces in $\mathcal{B}$ follow the same calculations.

Since we can easily obtain the ground truth vote for each predicted value ${n^j_i},d^j_i$, we directly supervise it by means of the $\mathcal{L}_1$ loss.
As for confidence $c^j_i$, similar to Eq.~\ref{basic_c}, we define the following loss term
\begin{equation}
    \mathcal{L}^{Basic}_{p_c} = \lambda_3\sum_{p_j \in P} \sum_{i \in \mathcal{B}} \left\|c^j_i - exp(\frac{|d^j_in^j_i-f^i_{j}(p_j)r^{gt}_i|}{-k_2})\right\|_1,
    \label{basic_pc}
\end{equation}
where $k_2$ is a constant hyper-parameter and $f^i_{j}(p_j)$ denotes the ground truth distance of $p_j$ to the face $i$ in $\mathcal{B}$.
Considering now again the top face $y+$, we accordingly obtain 
\begin{equation}
    f^{y+}_j({p_j}) = \frac{s^{gt}_{[y+]}}{2} - R_{gt}^T(p_j - t_{gt}),
\end{equation}
with $s_{[y+]}$ being the size along $y+$ direction, and $R_{gt}$ and $t_{gt}$ referring to the ground truth rotation and translation.

Noteworthy, the contributions of our confidence-weighted bounding box voting are two-fold. 
First, it enforces geometric consistency terms within the \textbf{PBP} stream. Secondly, it also enhances the learning of intra-category geometric features as experimentally proven in the evaluation section.
In general, the closer a point is to a bounding box plane, the more confidence it should have to accurately vote for the bounding box plane, as it is less affected by intra-class variations.
As a consequence, these geometric guidance enforces the learning of global (green) and per-point (blue) features in Fig.~\ref{pipeline} to capture more specific category-level traits.

\begin{figure}[t]
  \centering
  \includegraphics[width=0.99\linewidth]{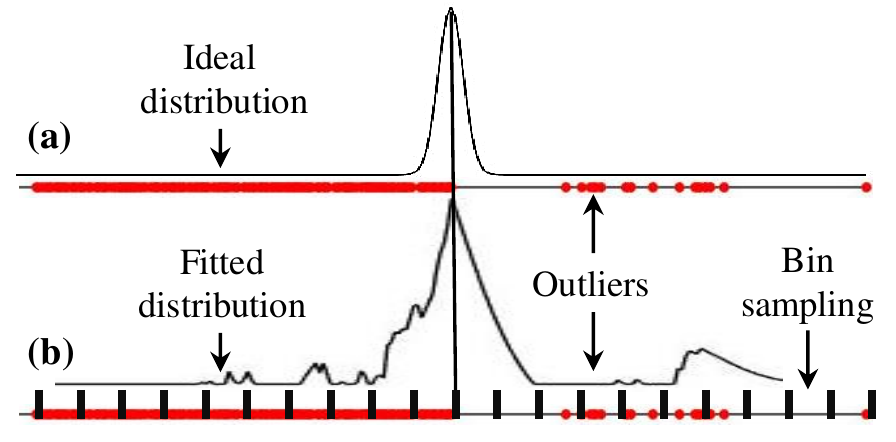}
  \caption{Illustration of how we predict $\{\bm{s}\}$ based on the point cloud only.
  Given all points projected along the y positive axis, we aim to approximate the ideal distribution of size $\{\bm{s}\}$ with combined exponential functions in Eq.~\ref{fs}, as shown in (b).
  }
  \label{outlier}
\end{figure}

\subsection{Point Cloud - Pose Geometric Consistency}
\label{PPGC}
While most category-level methods~\cite{fs-net, dualposenet, sgpa} predict the pose from uniformly sampled point clouds, they do not consider the geometric relation between pose and the point cloud explicitly. Instead they resort to individual loss terms for each pose parameter to learn such geometric relations implicitly~\cite{fs-net,dualposenet,sgpa}. 
On another note, it is not difficult to deduce that, when transforming the point cloud to the canonical view via the inverse of $\{\bm{R, t}\}$ and removing all outliers, the size $\{\bm{s}\}$ can be simply computed as the distance between the outermost points along each axis.
In this paper, we turn the above transformation into supervision terms to explicitly enforce these geometric relations and, thus, enhance performance.

To this end, we first transform the point cloud from the camera view to a predefined canonical view. Similar to the point matching loss used in instance-level pose prediction~\cite{GDRN, sopose, li2019deepim}, we then minimize
\begin{equation}
    \mathcal{L}_{(R, t)}^{PC} = \sum _{p \in P}\left\|(R^T (p - t) - p^{c})\right\|_1,
    \label{pm}
\end{equation}
with $P$ being the sampled point cloud and $p^{c}$ being the corresponding ground truth of $p$ under the canonical view.

Next, estimating the correct metric scale can be generally considered as the task of finding the 3D bounding box which encapsulates all object points whilst having minimal volume. 
Unfortunately, this is only true as long as we do not observe any outliers. 
Since this is rarely the case when dealing with real data, we instead relax this constraint and try to enforce having as \textit{many} points as possible in the bounding box of minimal volume. 
Given sampled points with a few outliers, we first conduct bin sampling to ensure that the points follow an approximate uniform distribution.
To this end, we segment the points into 64 bins. Further, whenever one bin overflows we uniformly sub-sample a fixed number of points. The final obtained point set is denoted as $P_s$. This process is also demonstrated in Fig.~\ref{outlier} (b).
Then given an estimate of $\{\bm{s}\}$, its probability to be the expected size is calculated as follows,
\begin{equation}
\begin{split}
    f(s) =& min(0, \frac{k_p}{|P_s|}f_p(s) \cdot f_d(s))
\end{split}
\label{fs}
\end{equation}
\begin{equation}
    f_p(s) =exp(\frac{k_n}{|P_s|}\sum_{p \in P_s} (\alpha(p, s) + 1))
\label{fps}
\end{equation}

\begin{equation}
    f_d(s) =\sum _{p \in P_s}\alpha(p, s)exp(-k_s|s-p|^2)),
\label{fds}
\end{equation}
where $\{k_s, k_n\}$ are constant hyper-parameters.
Further, $k_p$ is a normalization parameter and $\alpha(p, s) = 1$ if $p\leq s$, otherwise $\alpha(p, s) = -1$.
From this, our geometric loss term for the scale $\{\bm{s}\}$ can be derived as
\begin{equation}
  \mathcal{L}_{(s)}^{PC} = \sum_{i \in \mathcal{B}'} \left\|(1.0- f(s_i))\right\|_1,
\end{equation}
where $\mathcal{B}'=\{x+, y+, z+\}$.
Note that this term is only meaningful when the corresponding bounding box face in $\mathcal{B}$ is visible from the given viewpoint, hence, we turn off $\mathcal{L}_{(s)}^{PC}$ for occluded faces.

To summarize, our geometry-aware loss terms between the estimated pose and the corresponding object point cloud are defined as
\begin{equation}
  \mathcal{L}_{(R, t, s)}^{PC} = \lambda_4 \mathcal{L}_{(R, t)}^{PC} + \lambda_5\mathcal{L}_{(s)}^{PC},
\end{equation}
where $\lambda_*$ denote the balancing weights of the two terms.

\subsection{Bounding Box - Pose Geometric Consistency}
\label{BPGC}
From Sec.~\ref{BBoxVote}, we take $\{\bm{t, s}\}$ as the center and \{length, width, height\} of the bounding box from the point-wise voting.
Further, as aforementioned, we decompose the rotation $\{\bm{R}\}$ into its 3 columns $\bm{R}=[r_x, r_y, r_z]$, yet, only predict $r_x$ and $r_y$ together with the associated confidence, since they already fully describe the 3D rotation, \ie $r_z = r_x \times r_y$.
We have also shown that $r_x$ and $r_y$ correspond to the plane normals of the bounding box.
Thus given the estimated six bounding box faces $\{N_i, D_i\}$, $i \in \mathcal{B}$, we can define the following \textbf{Bounding box - Pose} consistencies
\begin{equation}
    \left\{\begin{matrix}
\mathcal{L}^{BB}_{(t)}= \sum_{i \in \{x, y, z\} }\left\||N_{i+}^Tt-D_{i+}|- |N_{i-}^Tt-D_{i-}|)\right\|_1\\ 
\mathcal{L}^{BB}_{(s)} = \sum_{i \in |\mathcal{B}|}\left\|{s_{i}- |N^T_{i}t-D_{i}|}\right\|_1 \\
\mathcal{L}^{BB}_{(R)} = \left\|r_y- N_{y+}\right\|_1 + \left\|{r_x}- N_{x+}\right\|_1\\ 
\end{matrix}\right.
\end{equation} 

The overall geometric consistency between pose and the estimated bounding box is then computed as the weighted sum of these individual terms
\begin{equation}
    \mathcal{L}^{BB}_{(R, t, s)} = \lambda_6 \mathcal{L}^{BB}_{(R)} + \lambda_7 \mathcal{L}^{BB}_{(t)} +  \lambda_8 \mathcal{L}^{BB}_{(s)}.
\end{equation}

\subsection{Overall Training Objective}
In summary, GPV-Pose employs the following loss function
\begin{equation}
    \mathcal{L} = \lambda_{Basic} \mathcal{L}^{Basic} + \lambda_{BB} \mathcal{L}^{BB}_{(R, t, s)} + \lambda_{PC} \mathcal{L}^{PC}_{(R, t, s)},
\end{equation}
where $\mathcal{L}^{Basic}$ contains all loss terms for fully supervising the learning of pose, rotation confidence, symmetrical reconstruction and point-wise bounding box voting.
Further, $\mathcal{L}^{BB}_{(R, t, s)}, \mathcal{L}^{PC}_{(R, t, s)}$ denote our geometric consistency terms as described in Sec.~\ref{PPGC} and Sec.~\ref{BPGC}, respectively.

%% file: sections/experiments1115.tex
\begin{table*}[ht]
\centering
\begin{tabular}{c|c|ccc|cccc|c}
\shline
Method & P.E. Setting & $3D_{25}$ & $3D_{50}$ & $3D_{75}$ &  $5^{\circ}2cm$& $5^{\circ}5cm$& $10^{\circ}5cm$& $10^{\circ}10cm$ & Speed(FPS)\\
\hline
NOCS~\cite{NOCS} & RGB-D & \textbf{84.9} & 80.5 & 30.1 & 7.2 & 10.0 & 25.2 & 26.7 & 5 \\
CASS~\cite{cass} & RGB-D & 84.2 & 77.7 & - & - & 23.5 & 58.0 & 58.3 & - \\
SPD~\cite{shape_deform} & RGB-D & 83.4 & 77.3 & 53.2 & 19.3 & 21.4 & 54.1 & - & 4 \\
CR-Net~\cite{cr-net} & RGB-D & - & 79.3 & 55.9 & 27.8 & 34.3 & 60.8 & - & - \\
SGPA~\cite{sgpa} & RGB-D & - & 80.1 & 61.9 & \textbf{35.9} & 39.6 & 70.7 & - & - \\
DualPoseNet~\cite{dualposenet} & RGB-D & - & 79.8 & 62.2  & 29.3 & 35.9 & 66.8 & - & 2 \\
\hline
DO-Net~\cite{donet} & D & - & 80.4 & \underline{63.7} & 24.1 & 34.8 & 67.4 & - & 10 \\
FS-Net~\cite{fs-net} & D & -& - & - & - & 28.2 & 60.8 & 64.6 & 20
\\
\hline
Ours(FS-Net) & D & $84.0$& $81.1$ & $52.0$ & $19.9$ & 33.9 & 69.1 & 71.0 & 20 \\
Ours(M) & D & 84.1 &  \underline{82.0} &  63.2 & 30.6 & \textbf{44.2} & \underline{72.1} & \underline{73.8} & 20 \\
Ours & D & \underline{84.2} &  \textbf{83.0} &  \textbf{64.4} & \underline{32.0} & \underline{42.9} & \textbf{73.3} & \textbf{74.6} & \textbf{20} \\
\shline
\end{tabular}
\vspace{-0.3cm}
\caption{\textbf{Comparison with state-of-the-art methods on REAL275 dataset.} Overall best results are in bold and the second best results are underlined.  
\textbf{P.E. Setting} lists the input data type for pose estimation.
Since FS-Net uses different detection results under 3DIoU, we reimplement it as \textbf{Ours(FS-Net)}.
Here \textbf{Ours(FS-Net)} inherits all loss terms of FS-Net but uses our pose decoder for fair comparison.
}
\label{tab_real275}
\end{table*}

\section{Experiments}
\textbf{Datasets.}
We evaluate GPV-Pose on the synthetic CAMERA25 and real REAL275~\cite{NOCS} benchmarks for category-level pose estimation and the real LineMod~\cite{Hinterstoisser2012a} dataset for instance-level pose estimation.
CAMERA25 contains 300K synthetic RGB-D images with rendered objects on top of virtual backgrounds, among which 25K images are withhold for testing.
The objects cover 6 categories, \ie \textit{bottle, bowl, camera, can, laptop} and \textit{mug}.
REAL275 is a more challenging real-world dataset with 13 different scenes. Thereby, 7 scenes with 4.3k images are provided for training, while the remaining 6 scenes with 2.7k images are employed for testing. 
REAL275 possesses the same categories as CAMERA25.
LineMod~\cite{Hinterstoisser2012a} is a widely-used instance-level 6D pose estimation benchmark, consisting of individual sequences with 13 objects undergoing mild occlusion.
We follow \cite{GDRN,li2019cdpn} and employ $\approx$15\% of the RGB-D images for training and the utilize the rest for testing.

\textbf{Implementation Details.}
We train our method purely on real data and follow~\cite{shape_deform,sgpa} to generate instance segmentation masks with an off-the-shelf object detector, \ie Mask-RCNN~\cite{maskrcnn}.
We uniformly sample 1028 points from the back-projected object depth map and feed them as input to GPV-Pose.
We employ several strategies for data augmentation including random scaling, random uniform noise, random rotational and translational perturbations, and bounding box based adjustment similar to FS-Net~\cite{fs-net}.
The parameters for all loss terms are kept unchanged during experimentation unless specified, with $\{\lambda_1, ...,\lambda_8, \lambda_{Basic}, \lambda_{BB}, \lambda_{PC}\} = \{1/8.0, 1/8.0, 1/8.0, 1.0, 1.0, 1.0, 1.0, 1.0, 8.0, 1.0, 1.0\}$.
We further set $\{k_1, k_2\}=\{13.7, 1/303.5\}$, $\{k_s, k_n\}=\{10, 0.5\}$ and keep $k_p$ fixed at 1.0, assuming an approximate point number of about 300 after bin sampling.
We employ the Ranger optimizer~\cite{ranger1,ranger2,ranger3} and run all our experiments on a single TITAN X GPU with batch size of 24 and base learning rate of 1e-4.
The learning rate is annealed at $72\%$ of the training phase using a cosine schedule.
The total training epoch is set to 70 for \textbf{Ours(M)}, in which we train a separate model for each category and 150 for \textbf{Ours} in which we train a single model for all categories.

\textbf{Evaluation Metrics.}
We follow~\cite{NOCS, sgpa, dualposenet} and report the mean precision of 3D intersection over union (IoU) at thresholds of 25$\%$, 50$\%$, 75$\%$ to jointly evaluate rotation, translation and size.
To directly compare errors in rotation and translation, we also adopt the $5^{\circ}2cm$, $5^{\circ}5cm$, $10^{\circ}5cm$, $10^{\circ}10cm$ metrics. 
A pose is thereby considered correct if the translation and rotation errors are both below the given thresholds.
For instance-level pose estimation task on LineMOD, we report the commonly employed ADD(-S) metric.

\subsection{Comparison with State-of-the-Art Methods}

\begin{table*}[t]
\centering
\scalebox{0.95}{
\begin{tabular}{c|ccc|c|ccc|cc|c|ccc}
\shline
\multirow{3}{0.3cm}{}&\multicolumn{3}{c|}{Network} & \multicolumn{6}{c|}{Loss Terms} & \multirow{3}{*}{$3D_{75}$} & \multirow{3}{*}{$5^{\circ}2cm$} & \multirow{3}{*}{$5^{\circ}5cm$} & \multirow{3}{*}{$10^{\circ}5cm$} \\
\cline{2-10}
& Conf. & B.Box & Symm. &\multirow{2}{*}{$\mathcal{L}^{Basic}$} & \multicolumn{3}{c|}{$\mathcal{L}^{BB}$} & \multicolumn{2}{c|}{$\mathcal{L}^{PC}$} & &  & & \\
\cline{6-10}
& Rot. & Voting & Recon. & &$\mathcal{L}^{BB}_{(R)}$ & $\mathcal{L}^{BB}_{(t)}$ & \multicolumn{1}{c|}{$\mathcal{L}^{BB}_{(s)}$} & $\mathcal{L}^{PC}_{(R, t)}$& $\mathcal{L}^{PC}_{(s)}$ & & & & \\
\hline
$A_1$&  &  &  & \checkmark & & & & & & 52.0 & 19.9 & 33.9 & 69.1 \\
$A_2$& \checkmark &  &  & \checkmark & & & & & & 56.9 & 22.7 & 35.7 & 70.0 \\
$A_3$& \checkmark &  & \checkmark & \checkmark & & & & & & 60.4 & 25.3 & 37.0 & 70.6 \\
\hline
$B_1$& \checkmark & \checkmark & \checkmark & \checkmark & \checkmark& & & & & 60.7 & 28.8 & 39.6 & 73.1 \\
$B_2$& \checkmark & \checkmark & \checkmark & \checkmark & &\checkmark & & & & 62.0 & 29.1 & 38.4 & 73.0 \\
$B_3$& \checkmark & \checkmark & \checkmark & \checkmark & & & \checkmark& & & 61.0 & 26.3 & 37.7 & 73.1 \\
$B_4$& \checkmark & \checkmark &  \checkmark& \checkmark &\checkmark & \checkmark& \checkmark& & & 63.2 & 29.9 & 39.7 & \textbf{73.3} \\
\hline
$C_1$& \checkmark &  & \checkmark & \checkmark & & & & \checkmark& & 61.5 & 27.4 & 40.3 & 73.0 \\
$C_2$& \checkmark &  & \checkmark & \checkmark & & & & \checkmark& \checkmark& 61.7 & 27.6 & 41.0 & 73.0 \\
\hline
$D_1$& \checkmark & \checkmark & \checkmark & \checkmark & \checkmark& \checkmark& \checkmark&\checkmark &\checkmark & \textbf{64.4} & \textbf{32.0} & \textbf{42.9} & \textbf{73.3}\\
\shline
\end{tabular} }
\vspace{-0.3cm}
\caption{\textbf{Ablation studies on different configurations of network architectures and loss terms on REAL275 datasets.}
\textbf{Conf. Rot.} refers to confidence-aware rotation representation. 
Without this term, we follow FS-Net to recover rotation matrix with SVD.
\textbf{B.Box Voting} refers to point-wise bounding box voting.
\textbf{Symm. Recon.} is symmetry-aware reconstruction.
Without this term, we directly reconstruct the input point cloud and $\mathcal{L}^{Basic}_{sym}$ is replaced by $\mathcal{L}^{Basic}_{recon} = \lambda_7 \zeta(P', P)$.
Overall best results are in bold.
}
\label{tab-abs}
\end{table*}

\begin{table}[t!]
\centering
\scalebox{0.99}{
\begin{tabular}{c|c|c|c}
\shline
Method & C.L. & ADD-(S) & speed(FPS) \\
\hline
PVNet~\cite{peng2019pvnet}  &  & 86.3 & 25 \\
G2L-Net~\cite{g2lnet}  &  & 98.7 & 23 \\
DenseFusion~\cite{densefusion}  &  & 94.3 & 16 \\
PVN3D~\cite{pvn3d}  &  & \textbf{99.4} & 5 \\
\hline
DualPoseNet~\cite{dualposenet}  & \checkmark & \textbf{98.2} & 2 \\
FS-Net~\cite{fs-net}  & \checkmark & 97.6 & 20 \\
Ours & \checkmark & \textbf{98.2} & 20 \\
\shline 
\end{tabular} }
\vspace{-0.3cm}
\caption{\textbf{Performance on LineMod.} 
We compare our method with competitors on the instance-level 6D pose estimation task on LineMod.
\textbf{C.L.} refers to category-level method.}
\label{tab-lm}
\end{table}

\begin{figure*}[t]
  \centering
  \includegraphics[width=0.99\linewidth]{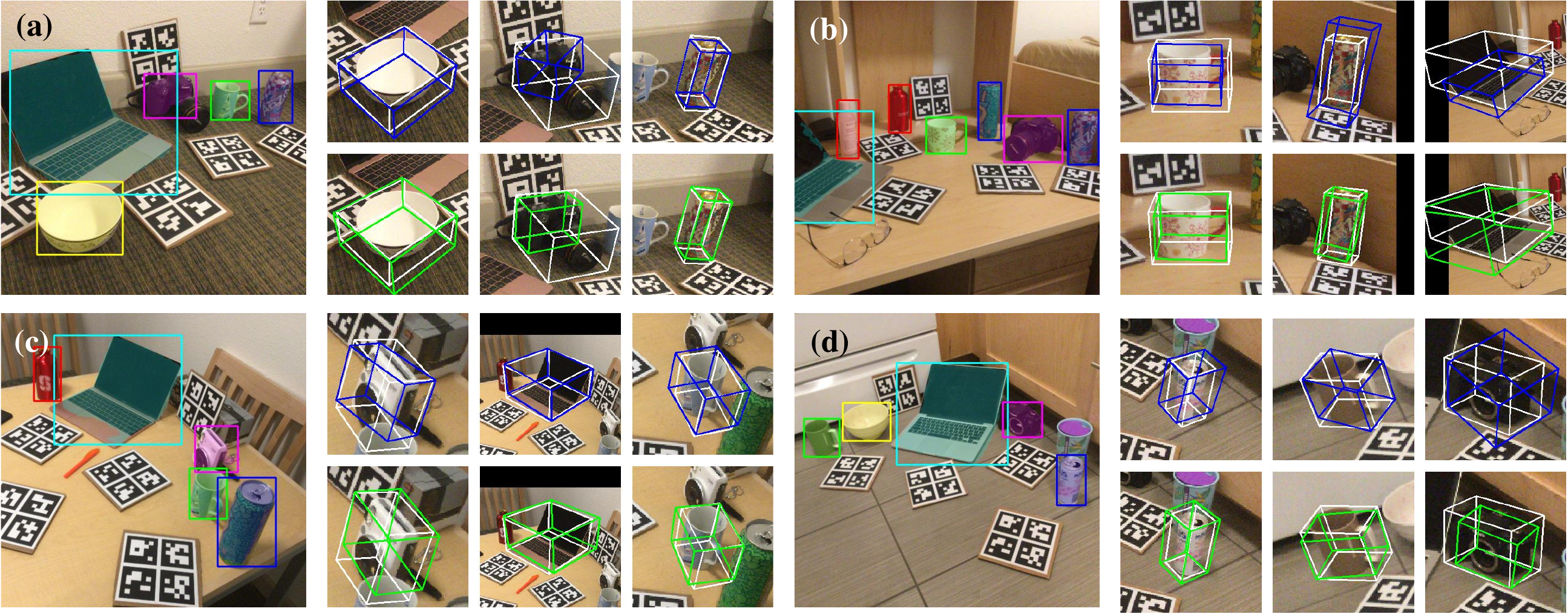}
  \caption{Qualitative results of our method (green line) and DualPoseNet (blue line). 
  Image (a)-(d) demonstrate 2D segmentation results.
  }
  \vspace{-0.5cm}
  \label{res_big}
\end{figure*}

\begin{figure}[t]
  \centering
  \includegraphics[width=0.99\linewidth]{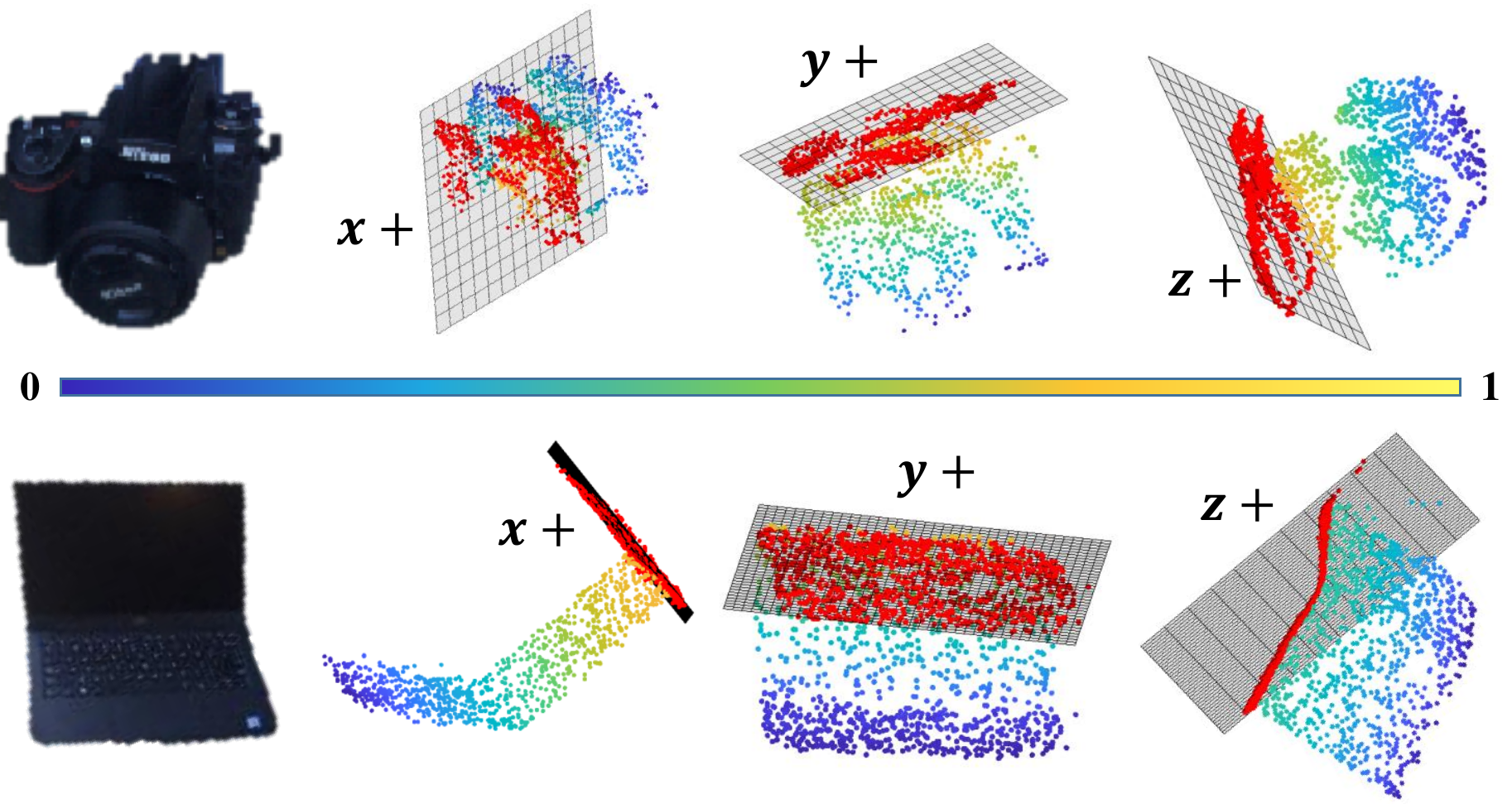}
  \caption{Demonstration of the point-wise bounding box voting results w.r.t three planes $\{x+, y+, z+\}$.
  GPV-Pose predicts point-wise directions, distances and confidences, moving the input points to the bounding box planes (the red points). Plane parameters are then calculated via confidence-weighted least squares.
  The color of each point (blue: low, yellow: right) reflects its confidence to support the target plane.
  We normalize the confidence for better visualization.
  In general, the closer a point is to the target plane, the higher its voting confidence for the plane.
  }
  \label{res_small}
\end{figure}

\begin{figure}[t]
  \centering
  \includegraphics[width=0.99\linewidth]{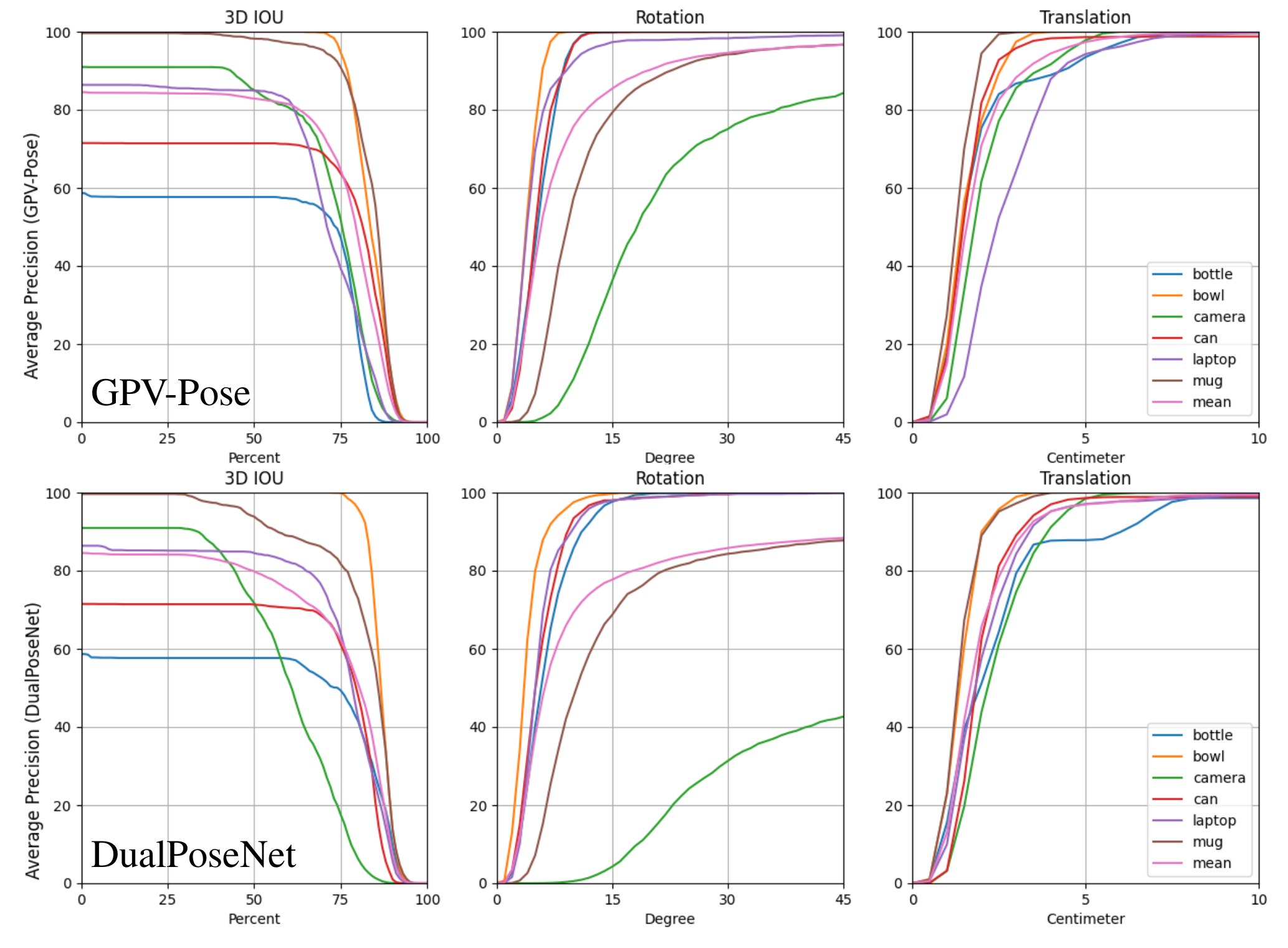}
  \caption{Per-category camparison between our method and DualPoseNet.
  We demonstrate average precision \vs different error thresholds on REAL275. 
  }
  \vspace{-0.3cm}
  \label{PDC_line}
\end{figure}

\textbf{Overall Performance on REAL275.}
In Tab.~\ref{tab_real275}, we compare GPV-Pose with state-of-the-art competitors on REAL275~\cite{NOCS}. 
In line with our work, also DO-Net~\cite{donet} and FS-Net~\cite{fs-net} both use only depth observations for pose estimation, whereas all other methods rely on RGB-D data during pose inference.
Notice that we provide three variants of our method.
\textbf{Ours(FS-Net)} inherits the loss terms of FS-Net but uses our pose estimation network architecture and serves as the baseline for our method.
\textbf{Ours(M)} trains a separate model for each category, while \textbf{Ours} trains a single model for all categories.
From Tab.~\ref{tab_real275}, it can be deduced that GPV-Pose achieves state-of-the-art performance w.r.t 5 out of total 7 metrics.
Specifically, for $3D_{50}$, GPV-Pose outperforms the previous best method NOCS with 83.0 \textit{vs.} 80.5. Furthermore, for the $5^{\circ}5cm$ metric, we surpass SGPA~\cite{sgpa} with 42.9 compared to 39.6, which increases by $3.3$.
Interestingly, the best performance for $5^{\circ}5cm$ is achieved by \textbf{Ours(M)} with 44.2.
In terms of $5^{\circ}2cm$, our method is inferior to SGPA~\cite{sgpa}.
The main reason may be that SGPA employs a point cloud based shape prior, while we only rely on the mean size of each category as prior.
As for our baseline method \textbf{Ours(FS-Net)}, we outperform it by a large margin w.r.t all metrics, clearly proving the effectiveness of our proposed contributions.
In Fig.~\ref{PDC_line}, we additionally present a detailed per-category comparison of our method with DualPoseNet~\cite{dualposenet}.
As one can easily deduce, we outperform DualPoseNet by a large margin, especially when viewing the 3D rotation results for non-symmetrical objects such as \textit{camera}. 
Moreover, Fig.~\ref{res_big} presents a qualitative comparison with DualPoseNet on REAL275.
GPV-Pose is capable of predicting accurate rotation and translation estimations, even when the objects are only partially detected, as shown in Fig.~\ref{res_big} (a) \textit{camera}, 
while DualPoseNet tends to fail on such challenging cases.
In addition, to get a better grasp of our methodology, in Fig.~\ref{res_small}, we demonstrate the bounding box voting results. It can be noted that the computed bounding box faces are accurate and the estimated confidence is reasonable (blue: low - yellow: high). 
Due to space limitations, we present our results on CAMERA25 in the supplementary material. In general, similar to REAL275, we again achieve comparable or superior performance compared to the state-of-the-art.

As speed is a vital factor for many applications, in the last column of Tab.~\ref{tab_real275}, we show the framerate of each method. 
Regardless of the object detection time, our method can infer at a real-time speed $>$50 FPS, and is thus very suitable for real applications with time requirements.
When using Yolo-V3~\cite{redmon2018yolov3} + ATSA~\cite{atsa} to extract object instances, the whole pipeline runs at about 20 FPS, faster than most other methods.

\textbf{Performance on Instance-level 6D Pose Estimation.}
We also apply GPV-Pose to the instance-level 6D pose estimation scenario and compare with state-of-the-art instance-level and category-level methods.
As shown in Tab.~\ref{tab-lm}, GPV-Pose achieves a comparable performance under the ADD(-S) metric (98.2\% for GPV-Pose compared to 98.2\% for DualPoseNet and 99.4\% for PVN3D~\cite{pvn3d}), whilst running almost in real time.

\subsection{Ablation Studies}

\textbf{Effect of Geometric Terms.} 
In Tab.~\ref{tab-abs} (B) and (C), we evaluate the performance w.r.t. different configurations of network architectures and geometric loss terms.
First, we gradually add each loss term (from $B_1$ to $B_4$) $\mathcal{L}^{BB}_{(R)}, \mathcal{L}^{BB}_{(t)}, \mathcal{L}^{BB}_{(s)}$, capturing the geometric relations between pose and the voted bounding box.
As can be observed, when adding these terms, the performance consistently improves from 60.4 to 63.2 referring to $3D_{75}$, and 25.3 to 29.9 w.r.t $5^{\circ}2cm$, which demonstrates the efficacy of the geometric consistencies.
Second, from $C_1$ to $C_2$ we evaluate the effectiveness of our loss terms $\mathcal{L}^{PC}_{(R, t)}$ and $\mathcal{L}^{PC}_{(s)}$, which are derived from the geometric relationship between pose and point cloud.
As before, the enforced consistency terms help again improve the estimator's performance (\ie we achieve an increase from 60.4 to 61.7 for $3D_{75}$).

\textbf{Effect of Confidence-Aware Rotation Prediction.}
When incorporating confidence into the rotation prediction, from $A_1$ to $A_2$ in Tab.~\ref{tab-abs}, the overall performance shows a significant leap forward from 52.0 to 56.9 referring to $3D_{75}$ and 19.9 to 22.7 for $5^{\circ}2cm$.
As aforementioned, for several categories one rotation vector is much easier to predict than the other and thus enables higher accuracy.
Exemplary, for \textit{laptop}, the average confidence for the $r_y$ normal lies around 0.88, whereas the average confidence for $r_x$ only amounts to 0.45. 
Since $r_y$ is typically perpendicular to the keyboard surface, it is a much easier target to estimate.
Therefore, exerting higher weight on $r_y$ benefits the performance.

%% file: sections/conclusion1022.tex
\section{Conclusion}
In this paper, we propose a novel category-level pose estimation network, which we dub GPV-Pose.
GVP-Pose jointly predicts the confidence-driven pose, symmetry-aware reconstruction and point-wise bounding box voting.
Two parallel streams of geometric consistencies, \textbf{Point Cloud - Pose} and \textbf{Point Cloud - Bounding Box - Pose}, are derived and converted into supervision terms to improve the pose accuracy.
GPV-Pose achieves superior performance on public datasets at a fast inference speed of 20 FPS, enabling real-time applications. 
In the future, we plan to incorporate GPV-Pose into holistic 3D understanding via first predicting the pose of each object and, secondly, describing the object interactions by means of a corresponding 3D scene graph.

%% file: main.bbl
\begin{thebibliography}{10}\itemsep=-1pt

\bibitem{cass}
Dengsheng Chen, Jun Li, Zheng Wang, and Kai Xu.
\newblock Learning canonical shape space for category-level 6d object pose and
  size estimation.
\newblock In {\em Proceedings of the IEEE/CVF conference on computer vision and
  pattern recognition}, pages 11973--11982, 2020.

\bibitem{sgpa}
Kai Chen and Qi Dou.
\newblock Sgpa: Structure-guided prior adaptation for category-level 6d object
  pose estimation.
\newblock In {\em Proceedings of the IEEE/CVF International Conference on
  Computer Vision}, pages 2773--2782, 2021.

\bibitem{g2lnet}
Wei Chen, Xi Jia, Hyung~Jin Chang, Jinming Duan, and Ales Leonardis.
\newblock G2l-net: Global to local network for real-time 6d pose estimation
  with embedding vector features.
\newblock In {\em Proceedings of the IEEE/CVF conference on computer vision and
  pattern recognition}, pages 4233--4242, 2020.

\bibitem{fs-net}
Wei Chen, Xi Jia, Hyung~Jin Chang, Jinming Duan, Shen Linlin, and Ales
  Leonardis.
\newblock Fs-net: Fast shape-based network for category-level 6d object pose
  estimation with decoupled rotation mechanism.
\newblock In {\em Proceedings of the IEEE/CVF Conference on Computer Vision and
  Pattern Recognition (CVPR)}, pages 1581--1590, June 2021.

\bibitem{robotics}
Xinke Deng, Yu Xiang, Arsalan Mousavian, Clemens Eppner, Timothy Bretl, and
  Dieter Fox.
\newblock Self-supervised 6d object pose estimation for robot manipulation.
\newblock In {\em 2020 IEEE International Conference on Robotics and Automation
  (ICRA)}, pages 3665--3671. IEEE, 2020.

\bibitem{sopose}
Yan Di, Fabian Manhardt, Gu Wang, Xiangyang Ji, Nassir Navab, and Federico
  Tombari.
\newblock So-pose: Exploiting self-occlusion for direct 6d pose estimation.
\newblock In {\em Proceedings of the IEEE/CVF International Conference on
  Computer Vision}, pages 12396--12405, 2021.

\bibitem{maskrcnn}
Kaiming He, Georgia Gkioxari, Piotr Doll{\'a}r, and Ross Girshick.
\newblock Mask r-cnn.
\newblock In {\em Proceedings of the IEEE international conference on computer
  vision}, pages 2961--2969, 2017.

\bibitem{FFB6D}
Yisheng He, Haibin Huang, Haoqiang Fan, Qifeng Chen, and Jian Sun.
\newblock Ffb6d: A full flow bidirectional fusion network for 6d pose
  estimation.
\newblock In {\em Proceedings of the IEEE/CVF Conference on Computer Vision and
  Pattern Recognition (CVPR)}, pages 3003--3013, June 2021.

\bibitem{he2020pvn3d}
Yisheng He, Wei Sun, Haibin Huang, Jianran Liu, Haoqiang Fan, and Jian Sun.
\newblock Pvn3d: A deep point-wise 3d keypoints voting network for 6dof pose
  estimation.
\newblock In {\em Proceedings of the IEEE/CVF conference on computer vision and
  pattern recognition}, pages 11632--11641, 2020.

\bibitem{pvn3d}
Yisheng He, Wei Sun, Haibin Huang, Jianran Liu, Haoqiang Fan, and Jian Sun.
\newblock Pvn3d: A deep point-wise 3d keypoints voting network for 6dof pose
  estimation.
\newblock In {\em Proceedings of the IEEE/CVF conference on computer vision and
  pattern recognition}, pages 11632--11641, 2020.

\bibitem{Hinterstoisser2012a}
Stefan Hinterstoisser, Cedric Cagniart, Slobodan Ilic, Peter Sturm, Nassir
  Navab, Pascal Fua, and Vincent Lepetit.
\newblock {Gradient Response Maps for Real-Time Detection of Textureless
  Objects}.
\newblock {\em TPAMI}, 2012.

\bibitem{hodan2020epos}
Tomas Hodan, Daniel Barath, and Jiri Matas.
\newblock Epos: Estimating 6d pose of objects with symmetries.
\newblock In {\em CVPR}, pages 11703--11712, 2020.

\bibitem{hu2020single}
Yinlin Hu, Pascal Fua, Wei Wang, and Mathieu Salzmann.
\newblock Single-stage 6d object pose estimation.
\newblock In {\em Proceedings of the IEEE/CVF Conference on Computer Vision and
  Pattern Recognition}, pages 2930--2939, 2020.

\bibitem{Kehl2017}
Wadim Kehl, Fabian Manhardt, Federico Tombari, Slobodan Ilic, and Nassir Navab.
\newblock Ssd-6d: Making rgb-based 3d detection and 6d pose estimation great
  again.
\newblock In {\em The IEEE International Conference on Computer Vision (ICCV)},
  Oct. 2017.

\bibitem{Kehl2016a}
Wadim Kehl, Fausto Milletari, Federico Tombari, Slobodan Ilic, and Nassir
  Navab.
\newblock {Deep Learning of Local RGB-D Patches for 3D Object Detection and 6D
  Pose Estimation}.
\newblock In {\em ECCV}, 2016.

\bibitem{labbe2020cosypose}
Yann Labb{\'e}, Justin Carpentier, Mathieu Aubry, and Josef Sivic.
\newblock Cosypose: Consistent multi-view multi-object 6d pose estimation.
\newblock In {\em European Conference on Computer Vision}, pages 574--591.
  Springer, 2020.

\bibitem{6drgbd}
Chi Li, Jin Bai, and Gregory~D Hager.
\newblock A unified framework for multi-view multi-class object pose
  estimation.
\newblock In {\em Proceedings of the european conference on computer vision
  (eccv)}, pages 254--269, 2018.

\bibitem{aticategory}
Xiaolong Li, He Wang, Li Yi, Leonidas~J Guibas, A~Lynn Abbott, and Shuran Song.
\newblock Category-level articulated object pose estimation.
\newblock In {\em Proceedings of the IEEE/CVF Conference on Computer Vision and
  Pattern Recognition}, pages 3706--3715, 2020.

\bibitem{li2019deepim}
Yi Li, Gu Wang, Xiangyang Ji, Yu Xiang, and Dieter Fox.
\newblock {DeepIM}: Deep iterative matching for 6d pose estimation.
\newblock {\em IJCV}, pages 1--22, 2019.

\bibitem{li2019cdpn}
Zhigang Li, Gu Wang, and Xiangyang Ji.
\newblock {CDPN}: {C}oordinates-{B}ased {D}isentangled {P}ose {N}etwork for
  {R}eal-{T}ime {RGB}-{B}ased 6-{DoF} {O}bject {P}ose {E}stimation.
\newblock In {\em ICCV}, pages 7678--7687, 2019.

\bibitem{donet}
Haitao Lin, Zichang Liu, Chilam Cheang, Lingwei Zhang, Yanwei Fu, and Xiangyang
  Xue.
\newblock Donet: Learning category-level 6d object pose and size estimation
  from depth observation.
\newblock {\em arXiv preprint arXiv:2106.14193}, 2021.

\bibitem{dualposenet}
Jiehong Lin, Zewei Wei, Zhihao Li, Songcen Xu, Kui Jia, and Yuanqing Li.
\newblock Dualposenet: Category-level 6d object pose and size estimation using
  dual pose network with refined learning of pose consistency.
\newblock {\em arXiv preprint arXiv:2103.06526}, 2021.

\bibitem{3DGC}
Zhi-Hao Lin, Sheng-Yu Huang, and Yu-Chiang~Frank Wang.
\newblock Convolution in the cloud: Learning deformable kernels in 3d graph
  convolution networks for point cloud analysis.
\newblock In {\em Proceedings of the IEEE/CVF Conference on Computer Vision and
  Pattern Recognition}, pages 1800--1809, 2020.

\bibitem{ranger1}
Liyuan Liu, Haoming Jiang, Pengcheng He, Weizhu Chen, Xiaodong Liu, Jianfeng
  Gao, and Jiawei Han.
\newblock On the variance of the adaptive learning rate and beyond.
\newblock In {\em International Conference on Learning Representations}, 2019.

\bibitem{lopez2011deformable}
Roberto~J L{\'o}pez-Sastre, Tinne Tuytelaars, and Silvio Savarese.
\newblock Deformable part models revisited: A performance evaluation for object
  category pose estimation.
\newblock In {\em 2011 IEEE International Conference on Computer Vision
  Workshops (ICCV Workshops)}, pages 1052--1059. IEEE, 2011.

\bibitem{manhardt2019explaining}
Fabian Manhardt, Diego~Martin Arroyo, Christian Rupprecht, Benjamin Busam,
  Tolga Birdal, Nassir Navab, and Federico Tombari.
\newblock Explaining the ambiguity of object detection and 6d pose from visual
  data.
\newblock In {\em Proceedings of the IEEE International Conference on Computer
  Vision}, pages 6841--6850, 2019.

\bibitem{manhardt2018deep}
Fabian Manhardt, Wadim Kehl, Nassir Navab, and Federico Tombari.
\newblock Deep model-based 6d pose refinement in rgb.
\newblock In {\em ECCV}, 2018.

\bibitem{manhardt2020cps}
Fabian Manhardt, Gu Wang, Benjamin Busam, Manuel Nickel, Sven Meier, Luca
  Minciullo, Xiangyang Ji, and Nassir Navab.
\newblock Cps++: Improving class-level 6d pose and shape estimation from
  monocular images with self-supervised learning.
\newblock {\em arXiv preprint arXiv:2003.05848v3}, 2020.

\bibitem{nie2020total3dunderstanding}
Yinyu Nie, Xiaoguang Han, Shihui Guo, Yujian Zheng, Jian Chang, and Jian~Jun
  Zhang.
\newblock Total3dunderstanding: Joint layout, object pose and mesh
  reconstruction for indoor scenes from a single image.
\newblock In {\em CVPR}, pages 55--64, 2020.

\bibitem{ozuysal2009pose}
Mustafa Ozuysal, Vincent Lepetit, and Pascal Fua.
\newblock Pose estimation for category specific multiview object localization.
\newblock In {\em 2009 IEEE Conference on Computer Vision and Pattern
  Recognition}, pages 778--785. IEEE, 2009.

\bibitem{park2019pix2pose}
Kiru Park, Timothy Patten, and Markus Vincze.
\newblock Pix2pose: Pixel-wise coordinate regression of objects for 6d pose
  estimation.
\newblock In {\em ICCV}, 2019.

\bibitem{peng2019pvnet}
Sida Peng, Yuan Liu, Qixing Huang, Xiaowei Zhou, and Hujun Bao.
\newblock Pvnet: Pixel-wise voting network for 6dof pose estimation.
\newblock In {\em CVPR}, 2019.

\bibitem{redmon2018yolov3}
Joseph Redmon and Ali Farhadi.
\newblock Yolov3: An incremental improvement.
\newblock {\em arXiv preprint arXiv:1804.02767}, 2018.

\bibitem{9d1}
Caner Sahin and Tae-Kyun Kim.
\newblock Category-level 6d object pose recovery in depth images.
\newblock In {\em Proceedings of the European Conference on Computer Vision
  (ECCV) Workshops}, pages 0--0, 2018.

\bibitem{savarese20073d}
Silvio Savarese and Li Fei-Fei.
\newblock 3d generic object categorization, localization and pose estimation.
\newblock In {\em 2007 IEEE 11th International Conference on Computer Vision},
  pages 1--8. IEEE, 2007.

\bibitem{hybridpose}
Chen Song, Jiaru Song, and Qixing Huang.
\newblock Hybridpose: 6d object pose estimation under hybrid representations.
\newblock In {\em Proceedings of the IEEE/CVF Conference on Computer Vision and
  Pattern Recognition}, pages 431--440, 2020.

\bibitem{arvr}
Yongzhi Su, Jason Rambach, Nareg Minaskan, Paul Lesur, Alain Pagani, and Didier
  Stricker.
\newblock Deep multi-state object pose estimation for augmented reality
  assembly.
\newblock In {\em 2019 IEEE International Symposium on Mixed and Augmented
  Reality Adjunct (ISMAR-Adjunct)}, pages 222--227. IEEE, 2019.

\bibitem{sundermeyer2020multi}
Martin Sundermeyer, Maximilian Durner, En~Yen Puang, Zoltan-Csaba Marton,
  Narunas Vaskevicius, Kai~O Arras, and Rudolph Triebel.
\newblock Multi-path learning for object pose estimation across domains.
\newblock In {\em CVPR}, pages 13916--13925, 2020.

\bibitem{Sundermeyer_2018_ECCV}
Martin Sundermeyer, Zoltan-Csaba Marton, Maximilian Durner, Manuel Brucker, and
  Rudolph Triebel.
\newblock Implicit 3d orientation learning for 6d object detection from rgb
  images.
\newblock In {\em ECCV}, 2018.

\bibitem{shape_deform}
Meng Tian, Marcelo~H Ang, and Gim~Hee Lee.
\newblock Shape prior deformation for categorical 6d object pose and size
  estimation.
\newblock In {\em European Conference on Computer Vision}, pages 530--546.
  Springer, 2020.

\bibitem{umeyama1991least}
Shinji Umeyama.
\newblock Least-squares estimation of transformation parameters between two
  point patterns.
\newblock {\em IEEE Transactions on Pattern Analysis \& Machine Intelligence},
  pages 376--380, 1991.

\bibitem{transformer}
Ashish Vaswani, Noam Shazeer, Niki Parmar, Jakob Uszkoreit, Llion Jones,
  Aidan~N Gomez, {\L}ukasz Kaiser, and Illia Polosukhin.
\newblock Attention is all you need.
\newblock In {\em Advances in neural information processing systems}, pages
  5998--6008, 2017.

\bibitem{6dpack}
Chen Wang, Roberto Mart{\'\i}n-Mart{\'\i}n, Danfei Xu, Jun Lv, Cewu Lu, Li
  Fei-Fei, Silvio Savarese, and Yuke Zhu.
\newblock 6-pack: Category-level 6d pose tracker with anchor-based keypoints.
\newblock In {\em 2020 IEEE International Conference on Robotics and Automation
  (ICRA)}, pages 10059--10066. IEEE, 2020.

\bibitem{densefusion}
Chen Wang, Danfei Xu, Yuke Zhu, Roberto Mart{\'\i}n-Mart{\'\i}n, Cewu Lu, Li
  Fei-Fei, and Silvio Savarese.
\newblock Densefusion: 6d object pose estimation by iterative dense fusion.
\newblock In {\em Proceedings of the IEEE/CVF conference on computer vision and
  pattern recognition}, pages 3343--3352, 2019.

\bibitem{wang2019densefusion}
Chen Wang, Danfei Xu, Yuke Zhu, Roberto Mart{\'\i}n-Mart{\'\i}n, Cewu Lu, Li
  Fei-Fei, and Silvio Savarese.
\newblock {DenseFusion}: 6d object pose estimation by iterative dense fusion.
\newblock In {\em CVPR}, pages 3343--3352, 2019.

\bibitem{GDRN}
Gu Wang, Fabian Manhardt, Federico Tombari, and Xiangyang Ji.
\newblock Gdr-net: Geometry-guided direct regression network for monocular 6d
  object pose estimation.
\newblock In {\em CVPR}, June 2021.

\bibitem{NOCS}
He Wang, Srinath Sridhar, Jingwei Huang, Julien Valentin, Shuran Song, and
  Leonidas~J Guibas.
\newblock Normalized object coordinate space for category-level 6d object pose
  and size estimation.
\newblock In {\em Proceedings of the IEEE/CVF Conference on Computer Vision and
  Pattern Recognition}, pages 2642--2651, 2019.

\bibitem{cr-net}
Jiaze Wang, Kai Chen, and Qi Dou.
\newblock Category-level 6d object pose estimation via cascaded relation and
  recurrent reconstruction networks.
\newblock {\em IEEE/RSJ International Conference on Intelligent Robots and
  Systems (IROS)}, 2021.

\bibitem{captra}
Yijia Weng, He Wang, Qiang Zhou, Yuzhe Qin, Yueqi Duan, Qingnan Fan, Baoquan
  Chen, Hao Su, and Leonidas~J Guibas.
\newblock Captra: Category-level pose tracking for rigid and articulated
  objects from point clouds.
\newblock {\em arXiv preprint arXiv:2104.03437}, 2021.

\bibitem{wohlhart2015learning}
Paul Wohlhart and Vincent Lepetit.
\newblock Learning descriptors for object recognition and 3d pose estimation.
\newblock In {\em CVPR}, 2015.

\bibitem{xiang2017posecnn}
Yu Xiang, Tanner Schmidt, Venkatraman Narayanan, and Dieter Fox.
\newblock {PoseCNN}: A convolutional neural network for 6{D} object pose
  estimation in cluttered scenes.
\newblock {\em RSS}, 2018.

\bibitem{ranger3}
Hongwei Yong, Jianqiang Huang, Xiansheng Hua, and Lei Zhang.
\newblock Gradient centralization: A new optimization technique for deep neural
  networks.
\newblock In {\em European Conference on Computer Vision}, pages 635--652.
  Springer, 2020.

\bibitem{zakharov2019dpod}
Sergey Zakharov, Ivan Shugurov, and Slobodan Ilic.
\newblock Dpod: Dense 6d pose object detector in rgb images.
\newblock In {\em ICCV}, 2019.

\bibitem{zhang2021holistic}
Cheng Zhang, Zhaopeng Cui, Yinda Zhang, Bing Zeng, Marc Pollefeys, and
  Shuaicheng Liu.
\newblock Holistic 3d scene understanding from a single image with implicit
  representation.
\newblock In {\em Proceedings of the IEEE/CVF Conference on Computer Vision and
  Pattern Recognition}, pages 8833--8842, 2021.

\bibitem{atsa}
Miao Zhang, Sun~Xiao Fei, Jie Liu, Shuang Xu, Yongri Piao, and Huchuan Lu.
\newblock Asymmetric two-stream architecture for accurate rgb-d saliency
  detection.
\newblock In {\em European Conference on Computer Vision}, pages 374--390.
  Springer, 2020.

\bibitem{ranger2}
Michael Zhang, James Lucas, Jimmy Ba, and Geoffrey~E Hinton.
\newblock Lookahead optimizer: k steps forward, 1 step back.
\newblock In H. Wallach, H. Larochelle, A. Beygelzimer, F. d\textquotesingle
  Alch\'{e}-Buc, E. Fox, and R. Garnett, editors, {\em Advances in Neural
  Information Processing Systems}, volume~32. Curran Associates, Inc., 2019.

\bibitem{zhou2019continuity}
Yi Zhou, Connelly Barnes, Jingwan Lu, Jimei Yang, and Hao Li.
\newblock On the continuity of rotation representations in neural networks.
\newblock In {\em Proceedings of the IEEE Conference on Computer Vision and
  Pattern Recognition}, pages 5745--5753, 2019.

\end{thebibliography}
